\documentclass[letterpaper, 10 pt, conference]{ieeeconf}

\IEEEoverridecommandlockouts

\usepackage[natbib=true,
            style=numeric,
            sorting=none,
            backend=biber,
            ]{biblatex}
\usepackage{common}
\usepackage{this_doc}

\title{\LARGE \bf
Prioritized Planning for Cooperative Range-Only Localization in Multi-Robot Networks
}

\author{Alan Papalia$^{1, 2}$, Nicole
Thumma$^{1}$, John Leonard$^{1}$
\thanks{This work was supported by ONR grant N00014-18-1-2832,
ONR MURI grant N00014-19-1-2571, and the MIT-Portugal
program.}
\thanks{$^{1}$Computer Science and Artificial Intelligence Lab (CSAIL),
Massachusetts Institute of Technology, 77 Massachusetts Ave, Cambridge MA 02139, USA
{\tt\small \{apapalia, thumman, jleonard\}@mit.edu}
}%
\thanks{$^{2}$Department of Applied Ocean Physics and Engineering, Woods Hole
Oceanographic Institution, 86 Water St, Woods Hole, MA 02543, USA} %
}

\begin{document}
\maketitle

\begin{abstract}
    We present a novel path-planning algorithm to reduce localization error for
    a network of robots cooperatively localizing via inter-robot range
    measurements. The quality of localization with range measurements depends on
    the configuration of the network, and poor configurations can cause
    substantial localization errors. To reduce the effect of network
    configuration on localization error for moving networks we consider various
    optimality measures of the Fisher information matrix (FIM), which have
    well-studied relationships with the localization error. In particular, we
    pose a trajectory planning problem with constraints on the FIM optimality
    measures.  By constraining these optimality measures we can control the
    statistical properties of the localization error. To efficiently generate
    trajectories which satisfy these FIM constraints we present a prioritized
    planner which leverages graph-based planning and unique properties of the
    range-only FIM.  We show results in simulated experiments that demonstrate
    the trajectories generated by our algorithm reduce worst-case localization
    error by up to 42\% in comparison to existing planning approaches and can
    scalably plan distance-efficient trajectories in complicated environments
    for large numbers of robots.
\end{abstract}
\section{Introduction}
\label{sec:intro}

Robotic systems have tremendous value in the advancement of scientific
exploration, with robots collecting data on planetary surfaces
\cite{Estlin2012}, outer space \cite{Gao2017science}, and in the deep oceans
\cite{Zhang2021science}. In these settings usage of multi-robot systems conveys
several advantages: observations can be made more rapidly, over larger areas,
and the robots can collaborate to more efficiently accomplish the mission.
However, a key challenge in real-world deployment of mobile robotic systems in
general is accurate localization.

\begin{figure}[!tbp] \centering \includegraphics[width=.9\linewidth,trim={0 0 0 4cm},clip
      ]{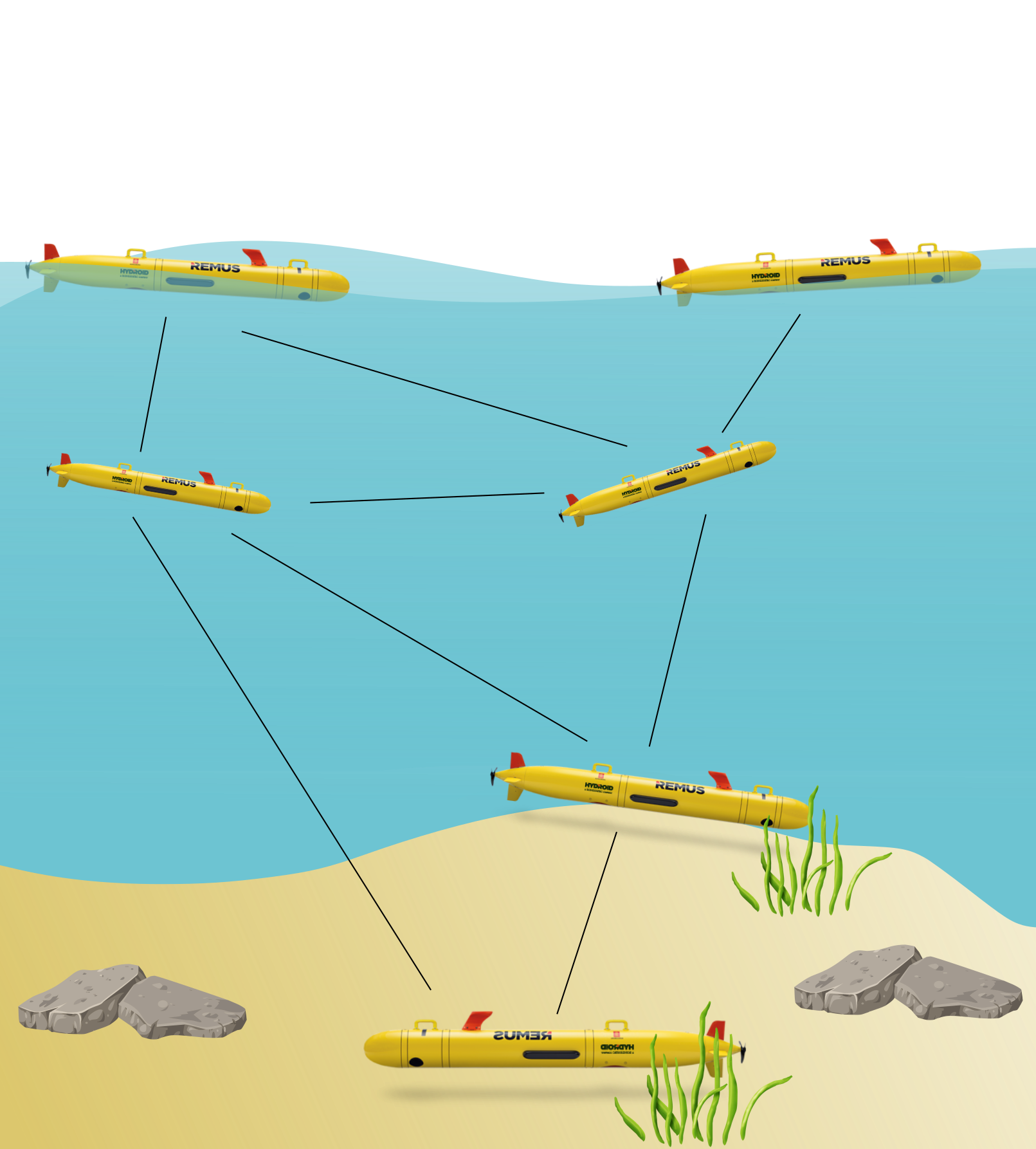} \label{fig:auv-network} \caption{Network of
            robots performing cooperative localization via range measurements. Edges
            between robots indicate inter-robot range measurements. Robots on the
            surface are assumed to have ground-truth positioning via GNSS.} \vspace*{-5mm}
\end{figure}

A common approach to multi-robot localization relies on teams in which the
robots use inter-robot ranging and a subset of the robots are considered anchors
\cite{Bahr12iros,Bahr07ecmr}. These anchors have high quality self-localization,
possibly due to global navigation satellite systems (GNSS) or high-precision
inertial systems, and enable for absolute localization of the remainder of the
network through the relative range measurements. However, the accuracy of
localizing the entire network via range measurements depends on the relative
positioning of the agents \cite{patwari2003relative} and the localization
problem can become ill-posed under certain network configurations, causing
localization techniques to fail drastically. As a result, as the
team of robots moves throughout space the localization of the non-anchor robots
can vary and substantially deteriorate. In safety-critical or high-cost missions
poor localization can lead to disastrous outcomes such as loss of expensive
equipment or even loss of life. To avoid the potential costs of poor
localization we propose a technique for planning trajectories for such robot
networks that effectively maintains a minimum localization quality at all
timesteps.

We measure the ability for a given configuration to be localized through
the Fisher information matrix (FIM) and FIM-based optimality measures from the
field of design of experiments. These measures are scalar values that relate the
FIM eigenvalues to various statistical properties of the localization problem
\cite{pukelsheim2006optimal}. By constraining these optimality measures with
lower bounds, the localization quality of the network can be controlled.
However, planning trajectories which constrain these measures is difficult, as
the optimality measures are complicated functions of the network configuration
which tightly couple the states of the separate robots in the network. The
primary challenge is in the inherent dimensionality of the planning problem, as
the individual robots cannot be easily decoupled and computational issues arise
with the resulting complexity of the problem. We present a framework for
decoupling the planning process and performing prioritized planning, i.e.
planning each robot individually in a predetermined sequence, while ensuring
that the network localization quality is maintained.

This paper extends our previous work on graph-based, prioritized, localizability-constrained planning \cite{papalia2020network} with the following contributions:

\begin{itemize}
      \item The relationship between localization quality and the theory of
            design of experiments is made clear.
      \item We present original proofs relating network connectivity, the FIM,
            and FIM optimality measures.
      \item This paper considers the case where there are sufficient anchor
            robots to fully constrain the localization problem, whereas
            \cite{papalia2020network} neglected the effects of anchor robots.
      \item More extensive experimental results with a more computationally
            efficient localizability-constrained planner and greater accuracy
            localization approach.
      \item The source code will be made freely available \footnote{\RepoURL}
\end{itemize}


\section{Related Work}
\label{sec:related-work}

As our work seeks to control robot networks to improve their localization we
first review similar works in the area of active localization. Then, as our
described problem is in multi-robot planning, we review relevant works in the
planning community. We note that our consideration of localization quality is
closely related to the well-known geometric dilution of precision
\cite{langley1999dilution}, but for brevity we do not discuss this in further
depth.

\subsection{Active Localization}

Prior works recognize the importance of geometry on the quality of range-only
localization and consider how to control members of a team to improve the
localization of others. Many approaches consider properties of either the FIM or
the closely related covariance matrix as measures of localization quality \cite{Bahr12iros,Walls2015iros, LeNy2018}. Existing approaches learn a policy
\cite{Chitre2010planning, tan2014cooperative} or perform belief space planning
\cite{Walls2015iros} for a single anchor vehicle to support a number of other
vehicles. A different approach developed a distributed algorithm for any number
of anchor vehicles to reduce the total uncertainty of any number of non-anchor
vehicles \cite{Bahr12iros}. However, these works assume that the non-anchor
vehicles have fixed trajectories and do not consider how the non-anchor
trajectories could be planned to improve their own localization.

Other works \cite{LeNy2018,Zelazo2015} present potential-field methodologies for
distributed trajectory planning for networks of anchor and non-anchor robots. In
\cite{LeNy2018} potential functions are presented for several FIM optimality
metrics. The work in \cite{LeNy2018} relates to earlier work in
\cite{Zelazo2015} which developed rigidity matrices%
\footnote{Under our assumptions of Gaussian or log-normal
    measurement distributions, the FIM we consider is a rigidity
    matrix and E-optimality is equivalent to the infinitesimal rigidity metric
    \cite{LeNy2018}.}
to control the infinitesimal rigidity of such a
network of robots, a necessary property for both control and localization in the
range-only case. However, the approaches in \cite{LeNy2018,Zelazo2015} are both
potential-field based techniques.  In these approaches, local minima in the
potential fields trap the planner and prevent the network from reaching the goal
locations. While our approach considers the same optimality measures as \cite{LeNy2018}, our
approach does not rely on the gradient of the optimality metrics and thus does
not suffer from the very common local minima presented in these potential-field
based approaches.

\subsection{Multi-Robot Planning}

Each additional robot in the network increases the dimensionality of the
planning problem. This leads to a challenge in the field of multi-robot
planning known as the curse of dimensionality, which refers to how the size of
the planning space scales exponentially with increasing dimensions. Prioritized
planning, in which the robots in a network plan their trajectories independently
in a predetermined sequence, has been proposed to reduce the effects of
dimensionality \cite{VanDenBerg2005iros}. In a similar vein,
\cite{Wagner2011iros} presents the idea of subdimensional expansion. With this
approach, trajectories are planned independently on a graph until interactions
occur between robots, at which point the planner considers the joint
configuration space of the interacting robots.  Other works in multi-agent
trajectory planning consider combinations of discrete and continuous
representations of space \cite{Park2020icra, Tordesillas2020mader}. They also
reduce the dimensionality by planning for subgroups of the robots at a time as
opposed to the whole team at once, using different strategies to handle
potential conflicts between robots.  While these notions of dimensionality
reduction are a key part of the approach we present, none of these approaches
can be immediately extended to our problem as the FIM construction enforces that
each robot needs to be considered as always interacting with all other robots.
\section{Problem Formulation}
\label{sec:problem-formulation}

We begin by establishing the difference between anchor and non-anchor robots in
our network. We present the FIM that will be referred to throughout this paper,
as originally derived in \cite{patwari2003relative}, as well as the different
optimality criteria of the FIM which our approach applies to. Finally, in this
section, we formally define the problem this paper considers.

As in \cite{patwari2003relative, papalia2020network}, we assume that range
measurements are either Gaussian or log-normally distributed. Additionally, we
assume a sensing horizon which prevents robots from obtaining range measurements
to each other if they are beyond a certain distance from each other.

\subsection{Terminology and Notation}

Throughout this paper we use the abbreviation LC to refer to a localizability
constraint (LC) and LC\textsubscript{SAT} to refer to a set of positions that
satisfy a group of localizability constraints, i.e the localizability constraint
satisfying (LC\textsubscript{SAT}) set. All indexing is zero-indexed, i.e. the
first row in a matrix is the $0$th row. We use two separate neighbor functions:
$\neighGraph(\cdot)$ acts on a set of nodes on a planning graph and returns the
union of all neighboring locations of all locations in the set, $\neigh(\cdot)$
acts on a robot in a network and returns all of the robots it has range
measurements to.

\subsection{Anchor and Non-Anchor Indexing}

Anchor robots are robots which are considered to have known absolute position.
In practice, this position information could be from sources such as GNSS or
high-precision inertial navigation systems. We order the robots such that the
anchors are grouped first and followed by the non-anchors, with no particular
ordering within these subgroups. For example, if we have $\nanch$ anchors and
$\nnon$ non-anchors the indices are $\{0, \dots, \nanch, \dots,
(\nnon+\nanch-1)\}$.

\subsection{Fisher Information Matrix}

We present the FIM in \cref{eq:fim} as derived in \cite{patwari2003relative}.
We begin by defining an undirected graph $\Graph = (\Vertices, \Edges)$ over the network of
robots where the edge $(i,j)$ exists if there is a range measurement between
robots $i$ and $j$. The position of robot $i$ is denoted $x_i \in \R^d$. The
FIM $F \in \R^{d \nnon \times d \nnon}$ is a ($d \times d$)-block-structured
matrix which is similar to a weighted graph Laplacian over $\Graph$ where the
Laplacian's rows and columns corresponding to the indices of anchor nodes have
been removed. The contributions of the sensor model in the FIM appear in the
following parameters: $\gamma$, which is equal to 1 if the sensor model is Gaussian and 2
if the model is log-normally distributed, and $\sigma$, the standard
deviation of the sensor distribution.

We first define the difference in robot positions and the distance between
robots as follows:
\begin{align}
    \DeltaIJ & \triangleq x_i - x_j \in \R^{d}       \label{eq:delta-ij} \\
    L        & \triangleq \| \DeltaIJ \|_{2} \in \R
\end{align}
From this, we define $b_{ij} \in \R^{d \nnon}$ and $q_{ij} \in \R^{d \nnon}$,
which relate the relative positions of robots in the network to sub-blocks of
the FIM. The vectors $b_{ij}$ and $q_{ij}$ are ($1 \times d$)-block-structured
vectors which relate nodes $i$ and $j$ whose $k$th-blocks are given in \cref{eq:b-ijk,eq:q-ijk}.
\begin{align}
    \label{eq:b-ijk}
    b_{ijk} & \triangleq \dfrac{\DeltaIJ}{\sigma L^{\gamma}}
    \begin{cases}
        1  & k = i            \\
        -1 & k = j            \\
        0  & \text{otherwise}
    \end{cases}                                \\
    \label{eq:q-ijk}
    q_{ijk} & \triangleq \dfrac{\DeltaIJ}{\sigma L^{\gamma}}
    \begin{cases}
        1 & k = i            \\
        0 & \text{otherwise}
    \end{cases}
\end{align}
We next define $\EdgesNA$ as the set of edges between a non-anchor node and an
anchor node and $\EdgesNN$ as the set of edges between two non-anchor nodes. In
the case of $\EdgesNA$ we assume without loss of generality that for each
edge-pair $(i,j) \in \EdgesNA$ the first value, $i$, refers to the non-anchor
node. With these edge sets the FIM, $F  \in \R^{d \nnon \times d \nnon}$, can be
expressed as follows:
\begin{equation}
    \label{eq:fim}
    F = \sum_{(i,j) \in \EdgesNA} b_{ij} b_{ij}^\top + \sum_{(i,j) \in \EdgesNN} q_{ij} q_{ij}^\top
\end{equation}
Note that there is no contribution from edges between two anchor nodes, as the
positions are known and thus no information is gained from such measurements.

As shown in \cite{LeNy2018,Zelazo2015}, the eigenvalues of the FIM are invariant
to rigid-motion transformations of the network members and the FIM is similar to
a weighted Laplacian of the connectivity graph of the network.  Unsurprisingly,
the weights in the Laplacian representation come exactly from the sensor
measurement model and the relative positioning of the robots with respect to
each other. By construction the FIM is a real, symmetric, positive semidefinite
matrix, meaning all eigenvalues are nonnegative.

\subsection{Localizability Criteria}
\label{subsec:optimality-criterion}

In design of experiments there are several common optimality measures which
relate the eigenvalues of the FIM to quality of the underlying estimation
\cite{pukelsheim2006optimal}. For our problem this can be thought of as
different measures of how the network configuration at a given timestep affects
the underlying information geometry and, as a result, the localization quality.

As our approach uses prioritized planning, these measures are observed with
increasing numbers of robots in the network, which corresponds to inference
problems of increasing dimensionality. Because of this changing dimensionality,
it can be difficult to relate the values of these optimality measures as robots
are added to the network. For this reason, we focus on A-optimality and
E-optimality in this paper, as they are relatively consistent as robots are
added, and do not consider other popular measures such as D-optimality, which
measures the volume of the covariance ellipsoid and thus grows geometrically
with increasing robots.%
\footnote{In addition, LCs could be designed which do not involve the FIM or
    information theory, but within this paper we strictly consider FIM optimality
    measures as LCs.}

We present A- and E-optimality in \cref{eq:a-optimality,eq:e-optimality} and
refer the interested reader to \cite{pukelsheim2006optimal} for theoretical
properties of these measures as well as a list of other possible measures. We
define $\bfp \in \R^{dn}$ as the vector of robot positions for $n$ robots in the
network and $F(\bfp)$ as the FIM that is formed from that configuration of
robots. The optimality measures are then as follows:
\begin{align}
     & \argmax_{p}~\AOptimalCost & (\text{A-optimal design}) \label{eq:a-optimality} \\
     & \argmax_{p}~\EOptimalCost & (\text{E-optimal design}) \label{eq:e-optimality}
\end{align}
We consider LCs to be predetermined, user-specified
minimum values on these measures. Under the LCs defined by
\cref{eq:a-optimality,eq:e-optimality} the set of allowed locations is
$\bf{p_{all}} = \{\bfp \in \R^{dn} : \ExampleConstraints\}$ for some
user-defined values of $\alpha$ and $\beta$. For a general set of such
user-defined criteria, we refer to network configurations which satisfy those
criteria as localizability-constrained.

\subsection{Planning Problem Definition}
\label{subsec:problem-description}

Given the ability to compute the FIM for a given network, we describe our
localizability constrained planning problem. This problem takes as input a set
of beginning locations ($\bfp (0)$), a set of goal locations ($\bfp (t_f)$), and
a set of LCs $\{\ExampleConstraints\}$. Given these inputs, the problem is to
find trajectories for each robot ($\bfp_{i} (t)$) such that each robot begins at
its starting position, ends at its goal position, and the LCs of the network are
satisfied at every timestep.

\section{Localizability Constrained Planning}
\label{sec:lcgp}

To solve the planning problem described in Section~
\ref{subsec:problem-description} we present a prioritized, graph-based planner
which plans for all robots on a single, common graph. To allow for scalability
to larger numbers of robots we use a prioritized approach
\cite{VanDenBerg2005iros} to reduce the computational cost of planning from the
$dn$-dimensional joint-configuration space of the robot network to the cost of
planning in the $d$-dimensional configuration space of just a single robot. The
graph-based planning is a probabilistic roadmap \cite{Kavraki1996} approach in
which we consider the network to move along the graph in discrete timesteps and
enforce LCs at each timestep. We make use of
constraint sets from our previous work \cite{papalia2020network} to efficiently
enforce the LCs. In particular, we require that the
trajectories planned for each robot $i$ stay within what we term the valid set,
$\valid_{i,t}$ in \cref{eq:valid-set}, for all timesteps $t$.

\subsection{Constraint Sets}
\label{subsec:constraint-sets}

The trajectory of the robot network can be thought of as a time-indexed sequence
of networks, with each network corresponding to a unique timestep. Similarly,
prioritized planning for each robot can be thought of as adding a single node to
each timestep-indexed network. This framework for viewing the problem is key in
understanding the benefit of the constraint sets described in this section, as
the constraint sets represent where nodes can be added to maintain certain
constraints on the network.

We plan each robot's trajectory separately and the localization
conditions for each timestep are independent of each other. For each robot $i$
and timestep $t$, there is a separate set of constraint sets $\{\reachable_{i,t}
    , \connected_{i,t} , \localizable_{i,t} , \valid_{i,t} \}$.
Constraint sets were first described in our previous work
\cite{papalia2020network} as a means to reduce the necessary computation to
perform planning. The constraint sets presented in this paper are the same
concept as those from our previous work, with the notable difference in that we
replace the rigid set, $\rigid_{i,t}$ in \cite{papalia2020network}, with the
localizability constraint satisfying LC\textsubscript{SAT} set, $\localizable_{i,t}$. In
\cite{papalia2020network} $\rigid_{i,t}$ was constraining the E-optimality of
the network, albeit without directly stating this relationship, whereas
$\localizable_{i,t}$ allows for a combination of such optimality constraints and
more directly ties into the theory of design of experiments.

Effectively, we use relationships between the constraint sets to reduce the
actual search space for each robot to a small subset of the actual nodes in the
planning graph. In this approach, the search space is reduced to what we term the
valid set, $\valid_{i,t}$, which, for non-anchor robots is the intersection of all
other constraint sets corresponding to robot $i$ at timestep $t$.

We begin by defining the boolean indicator function for the LC\textsubscript{SAT} set in
\cref{eq:lc-indicator}. We slightly abuse notation in using $\bfx$ to indicate
the position of an arbitrary node in the planning graph, whereas
$\mathbf{x}_{k,t}$ refers to the position of robot $k$ at timestep $t$ according
to a planned trajectory. The indicator function, $ \IndicatorLC (\mathbf{x}, i,
        t)$, evaluates to true if and only if at timestep $t$ the set of network
positions formed by the position $\bfx$ in addition to the positions of all
previously planned robots $(0, \dots, i-1)$ satisfies the user-determined
localizability criteria.

We formally define $\bfpx$ in \cref{eq:positions-it}, where $ \{
\mathbf{x}_{0,t}, \mathbf{x}_{1,t}, \dots, \mathbf{x}_{i-1,t} \} $ is a partial
assignment of robot positions from the robots before $i$ in the priority queue.
This arises from the prioritized planning, as the trajectories for robots $0$ to
$i-1$ will have been planned, so the positions of all of these robots at a given
timestep will be fixed. As such, $\bfpx$ suggests that the set of positions
considered is just indexed by an arbitrary position $x$ as all other positions
in the set are fixed given the timestep $t$.
\begin{align}
    \label{eq:positions-it}
    \bfpx & \triangleq \{\bfx\} \cup \{ \mathbf{x_{0,t}}, \mathbf{x_{1,t}},
    \dots, \mathbf{x_{i-1,t}} \}                                   \\
    \label{eq:A-condition}
    \A    & \triangleq (\AOptimalCostX \geq \alpha)                \\
    \label{eq:E-condition}
    \E    & \triangleq (\EOptimalCostX \geq \beta)
\end{align}
\begin{equation}
    \label{eq:lc-indicator}
    \IndicatorLC (\mathbf{x}, i, t)
    \triangleq \A
    \land \E
\end{equation}
The constraint sets are defined as follows, where $i$ and $j$ index robots, $t$
indexes timesteps, and $\rho$ is the radius of the sensing horizon:
\begin{align}
     & \reachable_{i,t} = \valid_{i,t-1}\cup~\neighGraph(\valid_{i,t-1})
     & (\text{Reachable Set})
    \label{eq:reachable-set}                                             \\
     & \connected_{i,t} = \bigcup_{j=1}^{i-1}
    \{ \bfx : \| \bfx - \mathbf{x}_{j,t} \|_{2} \leq \rho \}
     & (\text{Connected Set})
    \label{eq:connected-set}                                             \\
     & \localizable_{i,t} =
    \{ \bfx : \IndicatorLC ( \mathbf{x}, i, t ) \}
     & (\text{LC\textsubscript{SAT} Set})
    \label{eq:localizable-set}                                           \\
     & \valid_{i,t} =
    \begin{cases}
        \reachable_{i,t}                         & i \leq \nanch \\
        \reachable_{i,t} \cap \localizable_{i,t} & i > \nanch    \\
    \end{cases}
     & (\text{Valid Set})
    \label{eq:valid-set}
\end{align}
The reachable set $\reachable_{i,t}$ represents all points that robot $i$ can
get to by timestep $t$ while remaining within the valid set at all previous
timesteps. The connected set $\connected_{i,t}$ represents all points that robot
$i$ can occupy at timestep $t$ while being within the sensing horizon of one of
the positions of the already planned robots ($\mathbf{x}_{0,t},
    \ldots,\mathbf{x}_{i-1,t}$) at that timestep. The localizability-constrained set
$\localizable_{i,t}$ is the set of all positions which satisfy the described
localizability indicator function, defined in \cref{eq:lc-indicator}. Finally, the
valid set $\valid_{i,t}$ is the true set of interest in that it describes
locations which robot $i$ can reach by time $t$ while satisfying the
LCs at all times.

Within this work we assume the existence of $d+1$ anchor robots, which have
known position and therefore for these robots if a location is reachable we
consider it valid,%
\footnote{For certain applications one may consider imposing
    heuristic constraints on these vehicles to encourage them towards certain
    behaviors, e.g.  requiring they stay within a certain distance of each
    other, however in this work we do not explore such situations.}
whereas for the non-anchor robots a location must satisfy LCs as well as be
reachable to be valid. In fact, as we assume the existence of three anchors,
we prove in Lemma~\ref{lem:single-measure-singularity} and \cref{thm:localizability-req-connect} that if at least one non-anchor
robot has fewer than $d$ range measurements to it the FIM must be singular.%
\footnote{This can be intuitively understood, as for $d$-dimensional
localization $d$ range measurements are necessary to fully constrain the
position estimate.}

\begin{lemma}[$d$-Connectivity and FIM Singularity]
    \label{lem:single-measure-singularity}
    Given the FIM defined in \cref{eq:fim}, for any number of non-anchor nodes,
    if there exists a non-anchor node with less than $d$ neighbors the FIM will
    be singular.
\end{lemma}

\begin{theorem}[Localizability Requires $d$-Connectivity]
    \label{thm:localizability-req-connect}
    Consider the localizability constraint indicator $\IndicatorLC$ as in
    \cref{eq:lc-indicator}. We define a nontrivial constraint to be an
    assignment of $\{(\alpha, \beta) : (\alpha > -\infty) \lor (\beta >
        0)\}$.%
    \footnote{As the FIM is positive semidefinite, these
        lower bounds ($-\infty$, 0) are the minimum values of the respective
        images of \cref{eq:a-optimality,eq:e-optimality} and thus a 'trivial' assignment of these values
        satisfies $\IndicatorLC$ for all possible network configurations.}
    Given nontrivial constraints, if there exists $\bfx_i \in \bfpx$ such that
    $(|\neigh (\bfx_i) | < d)$ then the localizability constraints must be
    violated, i.e.  $\IndicatorLC$ must evaluate to false.
\end{theorem}

This relationship between the FIM and the connectivity of the non-anchor robots
leads to the result that for any reasonable LCs using any
of the criteria in \crefrange{eq:a-optimality}{eq:e-optimality} the
relationship in \cref{eq:localizable-subset-relationship} holds.
\begin{align}
     & \localizable_{i,t} \subseteq \connected_{i,t}
    \label{eq:localizable-subset-relationship}
\end{align}
We take advantage of the relationship in
\cref{eq:localizable-subset-relationship} when constructing the constraint sets,
as in general checking whether a given position is in the LC\textsubscript{SAT} set ($\bfx \in
    \localizable_{i,t}$) requires computing the eigenvalues of the FIM. As the FIM,
and thus its eigenvalues, differ for every possible $\bfx$, constructing the FIM
and finding eigenvalues becomes computationally demanding.  To reduce the
necessary computation, we only check states which are already connected, i.e.
$\bfx \in \connected_{i,t}$.
\begin{algorithm}
    \caption{Localizability-constrained priority planning over a shared graph for network of robots} \label{alg:planning}
    \begin{algorithmic}[1]
        \Procedure{Multi-Robot Planning}{}
        \State \(\text{trajectories} \gets \emptyset \)
        \State \(\valid_0 \gets  \text{Construct Valid Sets (0)}\)
        \State \(i\gets -\nanch \) \Comment{robot index}
        \State \(n\gets \) size of network
        \While{\(i < \nnon \)}
        \State \(\text{traj} \gets \text{Perform Planning}(\valid_i, i)\)
        \State \(\text{trajectories}_i \gets \text{traj}\)
        \State \(\valid_i=\text{Construct Valid Sets}(i)\)
        \If {$\valid_i=\emptyset$}
        \State \textbf{return} \(\emptyset\)
        \Comment{planning failed}
        \EndIf
        \State \(i \gets (i+1)\)
        \EndWhile
        \State \textbf{return} $\text{trajectories}$
        \Comment{planning successful}
        \EndProcedure
    \end{algorithmic}
\end{algorithm}
\begin{algorithm}
    \caption{Construct valid sets for robot $i$ at all timesteps from the
        trajectories of the previous robots} \label{alg:construct-valid}
    \begin{algorithmic}[1]
        \Function{Construct Valid Sets}{$i$}
        \State $P_{i,0} \gets $ robot $i$ start location
        \State $V_{i,0} \gets P_{i,0}$
        \State $t \gets 0$
        \State $x \gets $ goal location of robot $i$
        \While{$x \not\in \valid_{i,t}$}
        \State $N\gets$ all neighbors of $\valid_{i,t}$
        \State $\reachable_{i,t+1} \gets (\valid_{i,t}~\cup~N)$
        \If {$i\leq 2$}
        \State $\valid_{i,t+1} \gets \reachable_{i,t+1} $
        \Else
        \State $\localizable_{i, t+1}^{cand} \gets \reachable_{i,t+1} \cap
            \connected_{i,t+1}$ \label{line:localizable-candidate}
        \State $\valid_{i,t+1} \gets \text{Get LC\textsubscript{SAT} States}( \localizable_{i,
                t+1}^{cand}, i, t+1)$ \label{line:get-valid-set}
        \EndIf
        \State $t\gets t+1$
        \EndWhile
        \State \textbf{return} $V_i$ \{Returning valid sets\}
        \EndFunction
    \end{algorithmic}
\end{algorithm}
Furthermore, notice that the only set needed to perform planning is the valid
set, and as valid states must be reachable (from \cref{eq:valid-set}) we can
improve efficiency by only checking the LC\textsubscript{SAT} conditions for states which are also
reachable, i.e. $\bfx \in (\connected_{i,t} \cap \reachable_{i,t})$. This is
seen in \cref{line:localizable-candidate,line:get-valid-set} of
\cref{alg:construct-valid}, where the intersection of the reachable and
connected sets form the candidate states for the valid set.

\subsection{Prioritized Planning}

Given a planning graph, the prioritized planning proceeds by iterating over all
the robots in the network in the predetermined order, first generating the
constraint sets for that robot and then performing planning along the graph such
that the robot $i$ is within the valid set $\valid_{i,t}$ for all timesteps
$t$. This process is repeated for each robot in sequence until either all robot
trajectories have been planned or the planner has failed. The planner is
considered to fail if in the constraint set construction phase the valid sets
never contain the goal location for that robot. As after a given amount of
timesteps the valid sets will either become empty or reach a steady-state and no
longer include new positions, it is simple in practice to check if the
planner fails in this manner. If at some point the valid set $\valid_{i,t}$
possesses the goal location for robot $i$ then it is guaranteed by the
definition of the valid set that there exists a trajectory to that robot's
goal position which satisfies the LC\textsubscript{SAT} conditions at all timesteps. In this
event, set construction is considered successful.

In our previous work \cite{papalia2020network} we presented a conflict-based
approach for attempting to handle failures in the constraint set construction
phase. However, in practice we observed no occasions where the conflict
resolution approach led to successful replanning within any reasonable timespan.
For this reason we remove this aspect from the planning framework described here
in favor of an approach which fails outright when a trajectory
cannot be found. As the success of prioritized planning can depend on the
planning order, planning failures can be followed by a reordering and subsequent
replanning attempts.

As the anchor robots are necessary in evaluating the LCs,
the anchor robots are assumed to be planned before the non-anchor robots. Beyond
this, no requirements are made on the ordering of the planning, the method to
generate the planning graph, or the graph-based planner used.

\section{Results}
\label{sec:results}

\begin{table*}[!ht]
  \centering
  \caption{Information and results from experiments. Each test case
    represents a unique set of obstacles, goal locations, and number of robots
    in the network. The three environments we tested in have no obstacles, two
    small obstacles, and two large obstacles, which we have categorized as low,
    medium, and high environmental complexity. We present the planning time required, number of
    ordering attempts required to successfully generate a plan for (Orderings),
    the maximum and average localization errors (MLE and ALE, respectively), and
    the average distance a robot moved from the start to goal (AD). \\
    \footnotesize{*The average and max error for the RRT approach for test case
      2 are not included since one robot moved out of the sensing range of all
      other robots, which prevented the solver from producing a solution for
      the network during those timesteps.}
  }

  \begin{tabular}{| c | c | c | c | c | c | c | c | c |}
    \hline
    {\bf Test Case}    & {\bf \# Robots}     & {\bf Env. Complexity}   & {\bf Algorithm}   & {\bf Planning Time (s) } & {\bf Orderings} & {\bf ALE (m)}  & {\bf MLE (m)}  & {\bf AD (m)}    \\

    \hline
    \multirow{4}{*}{1} & \multirow{4}{*}{6}  & \multirow{4}{*}{Low}
                       & { LCGP (\bf Ours) } & 2.908                   & 1                 & 0.063                    & 0.129           & 36.272                                            \\
                       &                     &                         & { RRT }           & 0.381                    & --              & 0.050          & 0.132          & 42.257          \\
                       &                     &                         & { A* }            & 2.979                    & --              & 0.076          & 0.212          & 36.02           \\
                       &                     &                         & {Potential Field} & \textbf{0.160}           & --              & \textbf{0.049} & \textbf{0.087} & \textbf{34.641} \\
    \hline
    \multirow{4}{*}{2} & \multirow{4}{*}{8}  & \multirow{4}{*}{Medium}
                       & { LCGP (\bf Ours) } & 6.148                   & 7                 & \textbf{0.094}           & \textbf{0.199}  & 45.41                                             \\
                       &                     &                         & { RRT }           & \textbf{2.657}           & --              & *              & *              & 60.00           \\
                       &                     &                         & { A* }            & 3.968                    & --              & 0.103          & 0.213          & \textbf{45.00}  \\
                       &                     &                         & {Potential Field} & --                       & --              & --             & --             & --              \\
    \hline
    \multirow{3}{*}{3} & \multirow{4}{*}{8}  & \multirow{4}{*}{High}
                       & { LCGP (\bf Ours) } & \textbf{4.980}          & 1                 & \textbf{0.089}           & \textbf{0.516}  & 83.74                                             \\
                       &                     &                         & { RRT }           & 8.614                    & --              & 0.095          & 0.889          & 108.69          \\
                       &                     &                         & { A* }            & 5.017                    & --              & 0.171          & 0.705          & \textbf{80.64}  \\
                       &                     &                         & {Potential Field} & --                       & --              & --             & --             & --              \\

    \hline
    \multirow{3}{*}{4} & \multirow{4}{*}{12} & \multirow{4}{*}{High}
                       & { LCGP (\bf Ours) } & \textbf{7.711}          & 1                 & 0.073                    & \textbf{0.217}  & 83.55                                             \\
                       &                     &                         & { RRT }           & 13.846                   & --              & 0.091          & 0.377          & 111.54          \\
                       &                     &                         & { A* }            & 7.930                    & --              & \textbf{0.069} & 0.225          & \textbf{83.50}  \\
                       &                     &                         & {Potential Field} & --                       & --              & --             & --             & --              \\

    \hline
    \multirow{3}{*}{5} & \multirow{4}{*}{20} & \multirow{4}{*}{High}
                       & { LCGP (\bf Ours) } & 20.902                  & 7                 & 0.049                    & 0.119           & 78.38                                             \\
                       &                     &                         & { RRT }           & 20.147                   & --              & 0.050          & 0.136          & 100.20          \\
                       &                     &                         & { A* }            & \textbf{16.625}          & --              & \textbf{0.048} & \textbf{0.100} & \textbf{77.89}  \\
                       &                     &                         & {Potential Field} & --                       & --              & --             & --             & --              \\

    \hline
  \end{tabular}
  \label{tab:planning-results}
  \vspace*{-4mm}
\end{table*}

\begin{figure}[!tbp]
  \centering
  \includegraphics[height=4.5cm, width=.9\linewidth]{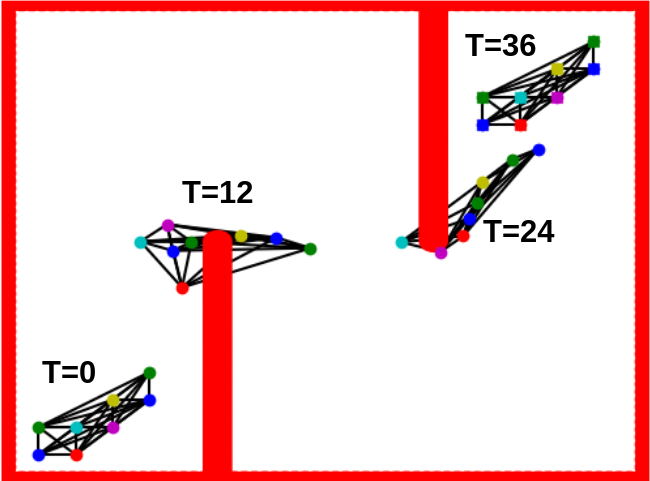}
  \caption{Snapshots of planned trajectory in example environment (Test Case 2).
    Dots indicate robots and lines indicate range measurements between robots.
    The configurations at timestep 0 and 36 are the start and goal. The network is qualitatively observed to
    maintain a triangulated configuration.}
  \label{fig:obstacle}
  \vspace*{-5mm}
\end{figure}

We tested our localization-constrained graph planning \textbf{(LCGP)} framework
over a number of two-dimensional simulated environments with varying numbers of
robots and obstacles. One of the tested environments and corresponding
trajectories in mid-execution is shown in Figure~\ref{fig:obstacle} for
reference. We set the first three robots as anchors (having known
position) and the remaining robots as non-anchors (having unknown positions).

We compare statistics on timing, planning, and localization for our planning
technique to three alternatives. We test against a prioritized RRT planner
\cite{kuffner2000rrt}, against a prioritized A* planner, and against the
potential field motion planner of \cite{LeNy2018} with E-optimality terms. All
planners were implemented in Python and run on an Intel i7-10875H processor.


To generate our planning graph, we use a Halton sequence to sample the
environment. Edges were made between every sampled node and neighbors
within 2 units, excluding edges that intersect obstacles. We
assume each robot to be a point robot and that there are no kinematic
restrictions on a robot's movement. To then perform planning on this graph, we
use heuristic-directed A* search with the Euclidean distance from a node to the
goal location as the heuristic.

Ordering is known to have substantial effects on the results of prioritized
planning. In the event of a failed planning, the planning order of the
non-anchor robots was reshuffled and planning was reattempted.  In cases 2
and 5, the initial ordering did not produce a viable trajectory, so an
alternative ordering was attempted.
\begin{figure}[!tbp]
  \centering
  \includegraphics[height=4.75cm, width=\linewidth,trim={0 3mm 0 1cm},clip]{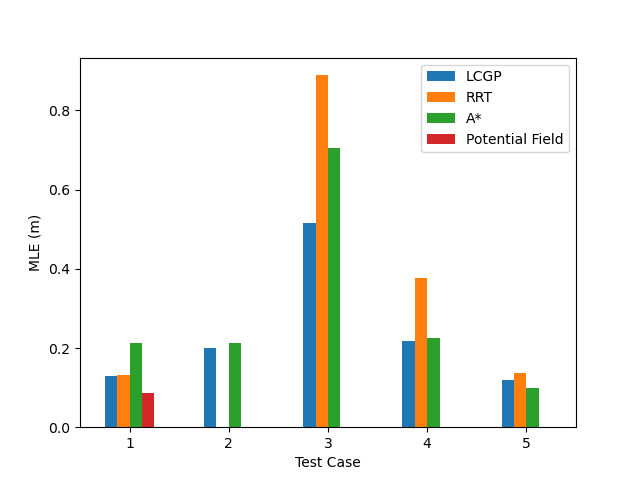}
  \caption{Max localization errors across all test cases.}
  \label{fig:roadmap}
  \vspace*{-5mm}
\end{figure}
During planning we constrained E-optimality to be greater than a value of 0.1
and placed no constraints on the A-optimality. This was justified by tests we
ran, which compared localization quality to various combinations of A-optimality
and E-optimality constraints and indicated that
A-optimality measures had relatively little effect on localization error
in comparison to E-optimality.

\subsection{Localization}

To compare the ability to localize over the course of a given path, we
implemented a nonlinear least-squares solver to find the maximum a posteriori
estimate and report both the mean error across all non-anchor robots at all
timesteps in the network and the max error of the mean across all non-anchor
robots at a single timestep. This error is calculated as the Euclidean distance
between the localization estimates and the ground truth positions.

The results for trajectories generated by our algorithm show the least max
localization errors in all instances except for the potential field in case 1
and A* in test case 5.  The greatest improvement was in test case 3, where LCGP
reduced the maximum localization error 42\% as compared to RRT and 26\% relative
to A*.  The average localization errors for our algorithm are also comparable or
lower than the alternatives for all cases. These results indicate that
localizability-constrained planning constructs trajectories which improve
range-only localization.

\subsection{Planning}
To evaluate the feasibility of each planner, we report the time
required to plan and the average distance travelled for each robot. In the case
of LCGP, the planning time includes the time required to build the planning
graph and replan using alternate orderings if necessary.

In environments with marked complexity or more challenging obstacles, notably
cases 3, 4, and 5, we found that our approach was as fast as or faster than
the RRT approach if the initial ordering succeeded. In test cases with any
obstacles (2-5), the potential field approach failed to find a trajectory to the
goal configuration due to local minima.  The relative speed of our LCGP planner
as compared to the RRT planner increased as either the number of robots or the
complexity of the environment increased.  However, in more simple environments
with a large amount of free space, the RRT planner was able to plan trajectories
up to an order of magnitude faster than the LCGP approach, as seen in cases
1 and 2.

Beyond planning time, we note that our LCGP approach found trajectories with
notably shorter execution distances compared to our RRT implementation. This
difference between our RRT implementation and our LCGP algorithm is likely due
to the use of A* to perform planning on the graph, which encourages the LCGP
planner to minimize travel distance. In comparison, the RRT approach would often
generate inefficient trajectories with no weight given to the travel time. We do
not compare the resulting distances to the theoretically optimal distances, but
these results along with qualitative inspection of the trajectories indicate
that the LCGP planner does generate distance-efficient trajectories.
\section{Conclusions}
\label{sec:conclusions}

In this paper we present a framework for efficiently planning
localizability-constrained trajectories for a network of robots localizing via
inter-robot ranging. We relate localization constraints to design of experiments
and prove that for $d$-dimensional localization every robot of unknown position
must have range measurements to at least $d$ other robots in the network to
satisfy our localization constraints. We validate the effectiveness of our
planning approach and its effect on the localization errors through extensive simulations.

Continuation of this work could consider the effects of different localizability
measures on the localization error. Additionally, future work could explore how
existing approaches in optimization-based trajectory planning can be modified to
accommodate localization constraints.

\section{Appendix}

\subsection{Proof of Lemma~\ref{lem:single-measure-singularity}}
\label{sec:proof:lem:single-measure-singularity}

To show that in a network where any non-anchor node has fewer than $d$ neighbors
the FIM is singular, we first demonstrate that the FIM has at least $d$
zero-eigenvalues when a non-anchor node has no neighbors and then demonstrate
that $d$ neighbors are the minimum number required to eliminate all of the
zero-eigenvalues. Our proof focuses on the column structure of the FIM, but the
FIM is symmetric, so the same arguments apply to the row-structure of the FIM.

Recall that the FIM $F \in \R^{dn \times dn}$ is a ($d \times
    d$)-block-structured matrix, with the $i$th ($dn \times d$) block-column
relating to position of non-anchor robot $i$. From \cref{eq:fim} it is
apparent that the FIM is the sum of rank-1 updates from the vectors $b_{ij}$ and
$q_{ij}$ corresponding to measurements in the network. It directly follows from
\cref{eq:b-ijk,eq:q-ijk} that if non-anchor node $i$ has no neighbors then the
$i$th block-column will be entirely zeros. By the size of the block-column, the
FIM must then have at least $d$ columns of zero-vectors and therefore have at
least $d$ zero eigenvalues.

By continuation of this idea, in the previous case the $d$ eigenvectors
corresponding to the zero eigenvalues are known to correspond to the $d$
dimensions in robot $i$'s position. The FIM is a series of rank-1 updates where
the dimensions affected by the updates are in the direction of the vectors
$b_{ij}$ and $q_{ij}$. Similarly, from \cref{eq:b-ijk,eq:q-ijk} it follows that
the direction of the update only affects the dimensions corresponding to robot
$i$'s position if the update corresponds to a measurement edge including robot
$i$.

As each rank-1 update can only affect a single direction in the $d$-dimensional
null-space, there must be $d$ linearly independent rank-1 updates to make the
FIM non-singular.%
\footnote{This requirement of $d$ linearly independent rank-1 updates to make
    the FIM non-singular relates to the property: $rank(A + B)
    \leq rank(A) + rank(B)$}
To have $d$ linearly independent rank-1 updates requires at
least $d$ measurement edges involving robot $i$. Therefore if robot $i$ has less
than $d$ neighbors the FIM is singular. \QED

\subsection{Proof of Theorem~\ref{thm:localizability-req-connect}}
\label{sec:proof:thm:localizability-req-connect}

For an arbitrary nontrivial constraint defined by $(\alpha, \beta )$ as in
\cref{thm:localizability-req-connect} and its corresponding constraint indicator
function $\IndicatorLC$. We show that in the event that any non-anchor node has
less than $d$ neighbors in the network any nontrivial constraint will be
violated. By Lemma~\ref{lem:single-measure-singularity}, if a
non-anchor node has less than $d$ neighbors the FIM will be singular. If the FIM
is singular it immediately follows that $\EOptimalCost = 0$ and thus any
nontrivial E-optimality constraint is violated. Similarly, as the FIM approaches
a singular representation the A-optimality metric approaches negative infinity,
i.e.  $\lim_{\lambda_{min} \to 0}~\AOptimalCost \to - \infty$, and thus in the
limit violates any nontrivial A-optimality constraint. As $\AOptimalCost$ is
undefined when the FIM is singular as the FIM cannot be inverted, we
consider the limiting behavior to show violation of nontrivial A-optimality
constraints. Thus if any non-anchor node has less than $d$ neighbors any
nontrivial constraint set will be violated. \QED

\printbibliography

@string{tra   = "{IEEE} Trans. Robotics and Automation"}

@string{ieee  = "Proc. of the IEEE"}

@string{joe   = "Journal of Oceanic Engineering"}

@string{icra  = "IEEE Intl. Conf. on Robotics and Automation (ICRA)"}

@string{iros  = "IEEE/RSJ Intl. Conf. on Intelligent Robots and Systems (IROS)"}

@string{ais = "Intl. Conf. on Autonomous and Intelligent Systems"}

@string{auv = "IEEE/OES Autonomous Underwater Vehicles Symposium"}

@inproceedings{Bahr07ecmr,
  author    = {A. Bahr and J. J. Leonard},
  year      = 2007,
  title     = {Minimizing Trilateration Errors in the Presence of
                  Noisy Landmarks},
  booktitle = {European Conference on Mobile Robots}
}

@inproceedings{Bahr12iros,
  author     = {A. Bahr and J.J. Leonard and A. Martinoli},
  fullauthor = {Alexander Bahr and John J. Leonard and A. Martinoli},
  title      = {Dynamic Positioning of Beacon Vehicles for Coop-
                  erative Underwater Navigation},
  booktitle  = iros,
  address    = {Algarve, Portugal},
  year       = 2012
}

@article{Chitre2010planning,
  author    = {Chitre, Mandar},
  doi       = {10.1109/AIS.2010.5547044},
  isbn      = {9781424471072},
  journal   = ais,
  publisher = {IEEE},
  title     = {Path Planning for Cooperative Underwater Range-Only Navigation using a Single Beacon},
  year      = 2010
}

@article{Estlin2012,
  author   = {Estlin, Tara A. and Bornstein, Benjamin J. and Gaines, Daniel M. and Anderson, Robert C. and Thompson, David R. and Burl, Michael and Casta{\~{n}}o, Rebecca and Judd, Michele},
  doi      = {10.1145/2168752.2168764},
  issn     = {21576904},
  journal  = {ACM Transactions on Intelligent Systems and Technology},
  number   = {3},
  title    = {AEGIS Automated Science Targeting for the MER Opportunity Rover},
  volume   = {3},
  year     = {2012}
}

@article{Gao2017science,
  author  = {Gao, Yang and Chien, Steve},
  doi     = {10.1126/scirobotics.aan5074},
  issn    = {24709476},
  journal = {Science Robotics},
  number  = {7},
  pmid    = {33157901},
  title   = {Review on space robotics: Toward top-level science through space exploration},
  volume  = {2},
  year    = {2017}
}

@article{Kavraki1996,
  author  = {Kavraki, Lydia E. and {\v{S}}vestka, Petr and Latombe, Jean Claude and Overmars, Mark H.},
  doi     = {10.1109/70.508439},
  issn    = {1042296X},
  journal = tra,
  number  = {4},
  pages   = {566--580},
  title   = {Probabilistic Roadmaps for Path Planning in High-Dimensional Configuration Spaces},
  volume  = {12},
  year    = {1996}
}

@inproceedings{kuffner2000rrt,
  title        = {RRT-connect: An efficient approach to single-query path planning},
  author       = {Kuffner, James J and LaValle, Steven M},
  booktitle    = icra,
  volume       = {2},
  pages        = {995--1001},
  year         = {2000},
  organization = {IEEE}
}

@article{langley1999dilution,
  title={Dilution of Precision},
  author={Langley, Richard B},
  journal={GPS world},
  volume={10},
  number={5},
  pages={52--59},
  year={1999}
}

@article{LeNy2018,
  author          = {{Le Ny}, Jerome and Chauvi{\`{e}}re, Simon},
  doi             = {10.23919/ACC.2018.8431528},
  isbn            = {9781538654286},
  issn            = {07431619},
  journal         = {Proceedings of the American Control Conference},
  pages           = {2788--2793},
  publisher       = {AACC},
  title           = {{Localizability-Constrained Deployment of Mobile Robotic Networks with Noisy Range Measurements}},
  volume          = {2018-June},
  year            = {2018}
}

@inproceedings{papalia2020network,
  title        = {Network Localization Based Planning for Autonomous Underwater Vehicles with Inter-Vehicle Ranging},
  author       = {Papalia, Alan and Leonard, John},
  booktitle    = auv,
  year         = 2020,
  organization = {IEEE}
}

@article{Park2020icra,
  arxivid = {1909.10219},
  author  = {Park, Jungwon and Kim, Junha and Jang, Inkyu and Kim, H. Jin},
  doi     = {10.1109/ICRA40945.2020.9197162},
  eprint  = {1909.10219},
  isbn    = {9781728173955},
  issn    = {10504729},
  journal = icra,
  pages   = {434--440},
  title   = {{Efficient Multi-Agent Trajectory Planning with Feasibility Guarantee using Relative Bernstein Polynomial}},
  year    = {2020}
}

@article{patwari2003relative,
  title     = {Relative location estimation in wireless sensor networks},
  author    = {Patwari, Neal and Hero, Alfred O and Perkins, Matt and Correal, Neiyer S and O'dea, Robert J},
  journal   = {IEEE Transactions on signal processing},
  volume    = {51},
  number    = {8},
  pages     = {2137--2148},
  year      = {2003},
  publisher = {IEEE}
}

@book{pukelsheim2006optimal,
  title     = {Optimal design of experiments},
  author    = {Pukelsheim, Friedrich},
  year      = {2006},
  publisher = {SIAM}
}

@article{tan2014cooperative,
  author    = {Tan, Yew Teck and Gao, Rui and Chitre, Mandar},
  doi       = {10.1109/JOE.2013.2296361},
  issn      = {03649059},
  journal   = joe,
  number    = {2},
  pages     = {371--385},
  publisher = {IEEE},
  title     = {Cooperative path planning for range-only localization using a single moving beacon},
  volume    = {39},
  year      = {2014}
}

@article{Tordesillas2020mader,
  archiveprefix = {arXiv},
  arxivid       = {2010.11061},
  author        = {Tordesillas, Jesus and How, Jonathan P.},
  eprint        = {2010.11061},
  pages         = {1--15},
  title         = {MADER: Trajectory Planner in Multi-Agent and Dynamic Environments},
  year          = {2020}
}

@article{VanDenBerg2005iros,
  author  = {{Van Den Berg}, Jur P. and Overmars, Mark H.},
  doi     = {10.1109/IROS.2005.1545306},
  isbn    = {0780389123},
  journal = iros,
  pages   = {2217--2222},
  title   = {Prioritized Motion Planning for Multiple Robots},
  year    = {2005}
}

@article{Wagner2011iros,
  author  = {Wagner, Glenn and Choset, Howie},
  doi     = {10.1007/978-3-642-33971-4_10},
  isbn    = {9783642339707},
  issn    = {18761100},
  journal = iros,
  pages   = {3260--3267},
  title   = {M*: A Complete Multirobot Path Planning Algorithm with Optimality Bounds},
  year    = {2011}
}

@article{Walls2015iros,
  author  = {Walls, Jeffrey M. and Chaves, Stephen M. and Galceran, Enric and Eustice, Ryan M.},
  doi     = {10.1109/IROS.2015.7353681},
  isbn    = {9781479999941},
  issn    = {21530866},
  journal = iros,
  pages   = {2264--2271},
  title   = {Belief Space Planning for Underwater Cooperative Localization},
  volume  = {2015-Decem},
  year    = {2015}
}

@article{Zelazo2015,
  archiveprefix   = {arXiv},
  arxivid         = {1309.0535},
  author          = {Zelazo, Daniel and Franchi, Antonio and B{\"{u}}lthoff, Heinrich H. and {Robuffo Giordano}, Paolo},
  doi             = {10.1177/0278364914546173},
  eprint          = {1309.0535},
  issn            = {17413176},
  journal         = {International Journal of Robotics Research},
  number          = {1},
  pages           = {105--128},
  title           = {{Decentralized rigidity maintenance control with range measurements for multi-robot systems}},
  volume          = {34},
  year            = {2015}
}

@article{Zhang2021science,
  author  = {Zhang, Yanwu and Ryan, John P. and Hobson, Brett W. and Kieft, Brian and Romano, Anna and Barone, Benedetto and Preston, Christina M. and Roman, Brent and Raanan, Ben Yair and Pargett, Douglas and Dugenne, Mathilde and White, Angelicque E. and Freitas, Fernanda Henderikx and Poulos, Steve and Wilson, Samuel T. and DeLong, Edward F. and Karl, David M. and Birch, James M. and Bellingham, James G. and Scholin, Christopher A.},
  doi     = {10.1126/SCIROBOTICS.ABB9138},
  issn    = {24709476},
  journal = {Science Robotics},
  number  = {50},
  pages   = {1--12},
  pmid    = {34043577},
  title   = {A system of coordinated autonomous robots for Lagrangian studies of microbes in the oceanic deep chlorophyll maximum},
  volume  = {6},
  year    = {2021}
}

\end{document}